\documentclass[conference]{IEEEtran}
\IEEEoverridecommandlockouts
\usepackage{cite}
\usepackage{amsmath,amssymb,amsfonts}
\usepackage{graphicx}
\usepackage{textcomp}

\usepackage[english]{babel}

\usepackage{mathtools}

\usepackage[noend,vlined,ruled]{algorithm2e}
\usepackage{bm}
\usepackage{amssymb} 
\usepackage{comment}
\usepackage{algcompatible}
\usepackage{ctable} 
\usepackage{xcolor}

\newcommand{\R}{\mathbb{R}}

\newcommand{\WK}{\underline{W}_K}
\newcommand{\WQ}{\underline{W}_Q}
\newcommand{\WV}{\underline{W}_V}

\newcommand{\W}{\underline{W}}
\newcommand{\K}{\underline{K}}
\newcommand{\V}{\underline{V}}
\newcommand{\Q}{\underline{Q}}

\newcommand\bred[1]{\textcolor{red}{\textbf{#1}}}
\newcommand\bgreen[1]{\textcolor{green}{\textbf{#1}}}

\usepackage{fancyhdr}

\fancypagestyle{firstpage}{%
  \fancyhf{} % clear headers/footers
  
  \fancyfoot[C]{\footnotesize
  Published in: \textit{ERK 2025 -- 34th Int. Electrotechnical and Computer Science Conf.}, 
  Portorož, Slovenia, 25–26 Sept. 2025. 
  Proceedings published by Društvo Slovenska sekcija IEEE, ISSN: 2591-0442 (online).}
}

\begin{document}

\title{A Neuro-Fuzzy System for Interpretable Long-Term Stock Market Forecasting
%{\footnotesize \textsuperscript{*}Note: Sub{-}titles are not captured in Xplore and should not be used}
\thanks{The authors acknowledge the financial support from the Slovenian Research and Innovation Agency -- ARIS (program funding number P2-0219).}
}
\author{\IEEEauthorblockN{ Miha Ožbot}
\IEEEauthorblockA{\textit{Faculty of Electrical Engineering} \\
\textit{University of Ljubljana}\\
Ljubljana, Slovenia \\
miha.ozbot@fe.uni{-}lj.si}%
\and
\IEEEauthorblockN{Igor Škrjanc}
\IEEEauthorblockA{\textit{Faculty of Electrical Engineering} \\
\textit{University of Ljubljana}\\
Ljubljana, Slovenia \\
igor.skrjanc@fe.uni{-}lj.si}%
\and
\IEEEauthorblockN{Vitomir Štruc}
\IEEEauthorblockA{\textit{Faculty of Electrical Engineering} \\
\textit{University of Ljubljana}\\
Ljubljana, Slovenia \\
vitomir.struc@fe.uni{-}lj.si}%
}%
%\and
% \IEEEauthorblockA{\textit{dept. name of %organization (of Aff.)} \\
% \and
%\and
%\IEEEauthorblockN{6\textsuperscript{th} Given Name %Surname}

\maketitle

\thispagestyle{firstpage}

\begin{abstract}
In the complex landscape of multivariate time series forecasting, achieving both accuracy and interpretability remains a significant challenge. This paper introduces the Fuzzy Transformer (Fuzzformer),  a novel recurrent neural network architecture combined with multi-head self-attention and fuzzy inference systems to analyze multivariate stock market data and conduct long-term time series forecasting. The method leverages LSTM networks and temporal attention to condense multivariate data into interpretable features suitable for fuzzy inference systems. The resulting architecture offers comparable forecasting performance to conventional models such as ARIMA and LSTM while providing meaningful information flow within the network. The method was examined on the real world stock market index S\&P500. Initial results show potential for interpretable forecasting and identify current performance tradeoffs, suggesting practical application in understanding and forecasting stock market behavior.
\end{abstract}
\section{Introduction}
\par\IEEEPARstart{I}{n} stock price forecasts, the interpretability of the model is desired to explain the predictions and extract semantic meaning that can be understood by investors. While black-box models achieve greater accuracy, their predictions are hard to explain, which is very important in economics. One of the most influential advancements in the field of deep neural networks is attention-based architectures; the connections of the input signals \cite{Vaswani2017} can be examined for technical analysis of the patterns from historical data. The interpretability of the relation between inputs and outputs was one of the cornerstones of fuzzy systems \cite{Xie2021}. We propose to combine recent advances in deep neural network time series forecasting \cite{Lim2021}, deep clustering \cite{Chang_2022}, and evolving fuzzy systems \cite{Ozbot_Lughofer_Skrjanc_2022} into a complex model, while maintaining enough transparency in the final layers to be interpreted by humans. 

\par Neuro-Fuzzy Systems (NFS) are models that can combine fuzzy logic rules as neurons in a network structure. Fuzzy and neuro-fuzzy models are commonly used for time series forecasting, specifically stock price prediction \cite{Pratama2014}. Examples include a Hammerstein-Wiener linguistic model \cite{Xie2021}, a fuzzy granular predictor for Bitcoin \cite{Garcia2019}, and an ensemble approach for S\&P500 prediction \cite{Jahandari2020}, though these typically focus on univariate or short-term forecasting.

\par Interpretability in fuzzy systems results from their structural separation between rule antecedents and functional or linguistic consequents \cite{Lughofer2013}. Antecedents are typically described using Gaussian clusters, obtained via unsupervised or supervised clustering. In this study, we use deep clustering to jointly learn the model with backpropagation. Various deep clustering approaches exist, such as autoencoders with k-means (e.g., DEKM \cite{Guo_Lin_Ye_2021}, FAE \cite{Aljalbout_Golkov_Siddiqui_Strobel_Cremers_2018}), or VAE-based generative models \cite{Chang_2022}. LSTM networks are a staple of sequence modeling in tasks such as regression and time series forecasting. In stock prediction, they are often combined with attention for long-term dependency modeling. Examples include the Temporal Fusion Transformer (TFT) \cite{Lim2021}, multi-horizon LSTM forecasting \cite{Aryal_Nadarajah_Rupasinghe_Jayawardena_Kasthurirathna_2020}, Indian stock market modeling \cite{Yadav2020}, NASDAQ-focused attention mechanisms \cite{Guo_Lin_Antulov-Fantulin_2019}, and transformer-only models \cite{Zeng_Chen_Zhang_Xu_2022}. Day-trading directional prediction with LSTM has also been explored \cite{Ghosh2022}. 
\par We propose a similar methodology by combining these advances with interpretable fuzzy systems. The main contributions of this paper are as follows:
%\begin{itemize}
{First, combining Long Short-Term Memory (LSTM) recurrent neural network architecture and the Multi-Head self-Attention (MHA) mechanism with Fuzzy Inference Systems (FIS) for multivariate multi-horizon time series forecasting (multi-step-ahead prediction).}
{ Second, an unsupervised approach to deep unsupervised multivariate Gaussian clustering of the low dimensional latent space.}
    %\item{ Online adaptation of the trained neuro-fuzzy network head, with evolving mechanisms.}
%\end{itemize}

\section{Methods} \label{sec:2}

\par Let \({\underline{X}}(k) {=} [\underline{x}(k{-}N), \ldots, \underline{x}(k)] {\in} \mathbb{R}^{N \times D_X}\) be the multivariate input data with \(D_X\) channels or input features, containing a main time series \(\underline{Y}(k) {=} [y(k{-}N), \ldots, y(k)] {\in} \mathbb{R}^{N}\), and other multivariate input data up to a discrete time step \(k\). Our objective is to predict the next \(H\) time steps of the main time series \(\underline{\hat{Y}}(k) {=} [\hat{y}(k{+}1), \ldots, \hat{y}(k{+}H)] {\in} \mathbb{R}^{H}\). 

  \begin{figure*}[!t]
\centering
  \includegraphics[width=0.9\linewidth]{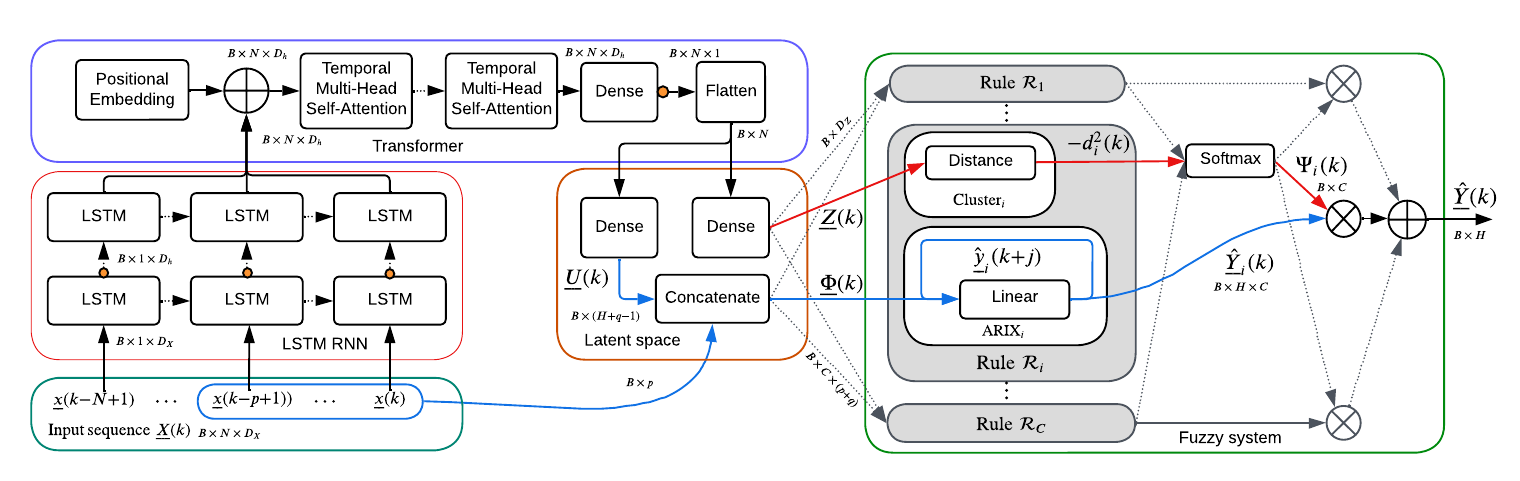}
    \caption{The Fuzzformer architecture. The orange dots at the outputs of the LSTM and Dense layers represent the dropout mechanisms. The red and blue connections illustrate the data flow of the antecedent latent space clustering samples and the local model regression variables, respectively. The variable $B$ is the batch size.\label{fig:1}
    }%
\end{figure*}

\par The proposed neuro-fuzzy transformer system, called the Fuzzformer, is a sequence-to-sequence model comprised of the following main layers: 1) An \textit{LSTM} network~\cite{Hochreiter_Schmidhuber_1997} to encode long-term dependencies between time steps; 2) A \textit{Multi-Head Self-Attention} network~\cite{Vaswani2017} for long-term information retention; 3) \textit{Fully connected layers} to reduce the encoded data into two low-dimensional latent space representations; 4) A \textit{fuzzy local model network} composed of multivariate Gaussian clusters~\cite{Ozbot_Lughofer_Skrjanc_2022} and ARIX local models~\cite{Prasad_Swidenbank_Hogg_1998} to generate the sequence forecast. The Fuzzformer architecture is presented in Figure~\ref{fig:1}.
%

%\subsection{Input Embedding and Spatial Attention}
%\par Input embedding is required to reshape the number of input features into a shape that can be distributed among the attention layer heads.
% \begin{equation}
% \begin{aligned}
% & i_t=\sigma\left(W_{i i} x_t+b_{i i}+W_{h i} h_{t-1}+b_{h i}\right) \\
% & f_t=\sigma\left(W_{i f} x_t+b_{i f}+W_{h f} h_{t-1}+b_{h f}\right) \\
% & g_t=\tanh \left(W_{i g} x_t+b_{i g}+W_{h g} h_{t-1}+b_{h g}\right) \\
% & o_t=\sigma\left(W_{i o} x_t+b_{i o}+W_{h o} h_{t-1}+b_{h o}\right) \\
% & c_t=f_t \odot c_{t-1}+i_t \odot g_t \\
% & h_t=o_t \odot \tanh \left(c_t\right)
% \end{aligned}
% \end{equation}

\subsection{Temporal Multi-Head Self-Attention}
\par The Multi-Head Attention (MHA) mechanism, as proposed in~\cite{Vaswani2017}, allows the model to focus on different parts of a sequence, by splitting the hidden features into several sub-spaces, computing an attention mechanism for each in parallel, and combining the outputs. Specifically, the input sequence $\underline{S} {=} [\underline{s}(1),...,\underline{s}(N)] {\in}\R^{N\times D_{in}}$ with length $N$ and input dimension $D_{in}$ is transformed with trainable parameter matrices $\WK {\in} \R^{D_{in}\times D_h}$, $\WQ {\in} \R^{D_{in}\times D_h}$, and $\WV {\in} \R^{D_{in}\times d_{out}}$ into keys $\K {=} \underline{S}\,\WK  {\in} \R^{N\times D_h}$, queries $\Q {=} \underline{S}\,\WQ {\in} \R^{N\times D_h}$, and values $ \V {=} \underline{S}\,\WV {\in} \R^{N\times d_{out}}$. A Scaled Dot-Product Attention is then computed as~\cite{Vaswani2017}:
\begin{equation}
   {A}(\Q,\K,\V) = \mathrm{softmax} \Big(\frac{\Q\K^\top} {\sqrt{D_h}}\Big) \V
\end{equation}
In multi-head attention, this is done multiple times in parallel, where each layer is called a head $h {=} 1,...,N_h$. The outputs of all heads are concatenated and reshaped with a linear transformation:
\begin{equation}
   {M} = [\underline{H}^1,...,{\underline{H}^{N_h}}] \W_O
\end{equation}
\begin{equation}
    \underline{H}^h= A (\underline{S}\,\W_\mathrm{q}^h,\underline{S}\,\W_K^h,\underline{S}\,\W_V) 
\end{equation}
where $\W_O {\in} \R^{(N_h d_{out})\times D_{out}}$ is the output projection. We used the same hidden dimensions as proposed in~\cite{Vaswani2017}, i.e, $D_h {=} d_{out} {=} D_{in}/N_h$. which results in the distribution of the $D_{in}$ features of the input sequence into $N_h$ sub-spaces.
%\par We implemented a shared parameter matrix $W_V$ for all heads based on the interpretable multi-head attention proposed in~\cite{Lim2021}. While this reduces the number of parameters, it allows for greater interpretability of the attention weights, i.e., the output is the same, just the attention of each head is different. Self-attention refers to the keys, queries, and values originating from the same input sequence $\underline{S}$. Additionally, the name temporal attention in this case refers to the attention being computed for the whole input sequence in order to find time-dependent connections between time steps as opposed to spatial attention, which learns relationships between features within a given time step.
%
\subsection{Fuzzy Inference System}\label{section:evolving_systems}
\par %The fuzzy Takagi-Sugeno model consists of multiple fuzzy rules with membership functions in the antecedent and an affine function in the consequent part of each rule.
A fuzzy Takagi-Sugeno rule with one membership function in the antecedent and an affine linear function in the consequent can be described as follows:
\begin{equation}
R_i :\quad \mathrm{IF} \quad \big(\underline{Z}(k) \sim  \mathcal{Z}_i \big) \quad \mathrm{THEN}  \quad \underline{\hat{Y}}_i(k),
\end{equation}
where $\sim$ denotes a soft membership to the fuzzy set $\mathcal{Z}_i$, $i{=}1,2,...,C$ is the index of the fuzzy rules, $j {=} 1,...,H$ is the future time step, $\underline{Z}(k) {\in}\R^{D_Z} $ is the antecedent input vector, $\underline{\hat{Y}}_i(k) {=} [\hat{y}_i({k {+}1}),...,\hat{y}_i(k{+}H)]$ is the output of the fuzzy rule.
The antecedent fuzzy sets $\mathcal{Z}_i$ can be represented by multivariate Gaussian clusters that can be rotated arbitrarily to define correlations between variables, since they can approximate a variety of data distributions~\cite{Skrjanc2020}
\begin{equation} 
d_i^2(k) =(\underline{Z}(k) {-}\underline{\mu}_i)^{\top}\underline{\Sigma}_i^{-1}(\underline{Z}(k){-}\underline{\mu}_i),
\end{equation}
where $d_i^2(k)$ is the Mahalanobis distance, $\underline{\mu}_i {\in}\R^{D_Z} $ and $\underline{\Sigma}_i {\in}\R^{D_Z \times D_Z}$ are the center and the covariance matrix of the cluster associated with the fuzzy rule $\mathcal{R}_i$. The membership functions are normalized with a softmax function \begin{math}\Psi_{i}(k) {=} \mathrm{e}^{-d_i^2(k)}/\sum_i^C\mathrm{e}^{-d_i^2(k)}{\in}[0,1]\end{math}, so that a unit partition is obtained \begin{math}\sum_{i=1}^c\Psi_{ki} {=} 1\end{math}. Finally, the output of the neuro-fuzzy model is aggregated from all activated rules as \begin{math}\underline{\hat{Y}}(k){=}\sum_{i{=}1}^C\Psi_{i}(k) \underline{\hat{Y}}_{i}(k)\end{math}.

\par In our case, we used the neural network encoder to generate the (non-)exogenous inputs $\underline{U}(k) {=} [u_1,...,u_p] {\in}\R^{p}$ for the ARIX model and the latent space vector $\underline{Z}(k)$ for the antecedent membership functions. The forecast of the ARIX local model of the rule $\mathcal{R}_i$ for the sequence $\underline{Y}(k)$ is then defined as:
%\begin{equation}
%\hat{y}_i(k+j)= (1-{A}_i(\mathrm{q}^{-1})) %\Delta^d {y}(k+j)+{B}_i(\mathrm{q}^{-1}) {u}(k+j)
%\end{equation}
\begin{equation}
\hat{y}(k + j) = \left(1 - q^{-1} \right)^{-d} \frac{B(q^{-1})}{A(q^{-1})} u(k + j)
\end{equation}
% \begin{equation}
% A(q^{-1}) (1 - q^{-1})^d y(k) = B(q^{-1}) u(k)
% \end{equation}
where $A_i\left(\mathrm{q}^{-1}\right) {=} 1 {+} a_1 \mathrm{q}^{-1} {+} ... {+} a_p \mathrm{q}^{-p}$ is the auto-regressive (AR) polynomial, $B_i\left(\mathrm{q}^{-1}\right) {=} b_1 \mathrm{q}^{-1} {+} ... {+} b_q \mathrm{q}^{-q}$ is the exogenous polynomial, $\mathrm{q}^{-1}$ is a discrete delay operator, i.e. ${y(k{-}p)} {=} \mathrm{q}^{{-}p}y(k)$, $p$ is the order of the auto-regressive model and $q$ is the order of the exogenous input, and $d$ is the order of integration. 
The model starts by using the known input time series values $\underline{X}(k)$ and then adds the recursively computed predictions $y_i(k+j)$ in a sliding-window way for all $j$, resulting in a combined forecast $\underline{\hat{Y}}_i(k) {=} [\hat{y}_i({k {+}1}),...,\hat{y}_i(k{+}H)]$.
\subsection{Training losses}
%The loss functions used in the proposed multi-task neural network can be separated into two groups: the forecast accuracy loss and the latent space clustering losses. One of the fuzzy layer rules (local model and clusters) may be more sensitive to changes in the shared bottom layers, and this imbalance can cause the overall model to perform poorly. A careful balance of the loss functions is required to maintain high accuracy and interpetability at the same time. 
\par To train the multi-horizon time series forecast, we employ the common Mean Squared Error (MSE) loss function, defined as: 
\begin{equation}
    \mathcal{L}_{\mathrm{MSE}} = \sum_k{||\underline{Y}(k)-\hat{\underline{Y}}(k)||^2},
\end{equation}
We used a "winner-takes-all" local optimization for the consequent local models to improve the interpretability of each local model, as opposed to the standard global optimization. This was done by computing the forward pass during training only for the rule with the highest activation, which forces each rule to have a good output locally. %Conversely, the intended fuzzy output was used to validate and test the model.
\par The encoder neural network generates latent features that are then clustered in an unsupervised way with multivariate Gaussian clusters. The Fuzzy C-means (FCM) clustering loss was used for deep unsupervised clustering:
\begin{equation}
    \mathcal{L}_{\mathrm{FCM}} = \sum_i^C\sum_k \mathrm{softmax}({-d_i^2(k)}) ||\underline{Z}(k) - \underline{\mu}_i||^2,
\end{equation}
%\par Two clustering regularization losses were used in the proposed approach. These losses improve the stability of the deep clustering method and also increase the interpretability of the model. 
Using only a clustering loss does not ensure separation of the clusters, as all clusters may converge to similar mean values and start to overlap. In order to keep the clusters distinguishable, a regularization loss is used to ensure cluster separation. We formulate an overlapping regularization loss as:
\begin{equation}\label{eq:loss_B}
    \mathcal{L}_O = \sum_m^C\sum_{n\neq m}^C\frac{1}{d_B(m,n)}
\end{equation} 
with the Bhattacharyya distance \cite{Lughofer2013}:
\begin{equation}\label{eq:bhattacharyya_distance}
\begin{split}
    d_B(m,n) =\, & \frac{1}{8} \big(\underline{\mu}_m{-} \underline{\mu}_n\big)^{\!\top}\bigg(\frac{\underline{\Sigma}_{m} + \underline{\Sigma}_{n}}{2}\bigg)^{\!\!-1}\! \big(\underline{\mu}_m{-} \underline{\mu}_n\big)+ \\ &
    \frac{1}{2}\ln\Bigg( \frac{\mathrm{det}\big(\frac{1}{2}(\underline{\Sigma}_{m} {+} \underline{\Sigma}_{n})\big)}{\sqrt{\mathrm{det}\underline{\Sigma}_m\mathrm{det}\underline{\Sigma}_n}}\Bigg),
    \end{split}
\end{equation}
where $m{=}1,2,...,C$ and $n{=}1,2,...,C$ are the indexes of the two compared clusters for $m{\neq}n$, and $d_B(m,n){=}d_B(n,m)$. 
%
%Importantly, we detached the cluster variances $\Sigma_i$ from the gradient computation, so that only the means $\mu_i$ are affected. This is because the Bhattacharyya distance returns a high value, i.e., the clusters are dissimilar or not overlapping, if the multivariate Gaussian distributions that represent the clusters have very dissimilar principal components in orientation and size. This can result in overlapping clusters with dissimilar orientations and sizes but similar centers. Furthermore, the Bhattacharyya distance \eqref{eq:bhattacharyya_distance} is saturated in the range $[10^{-3},3]$. The lower boundary is due to the inverse of the distance in the loss function \eqref{eq:loss_B}, and the upper boundary is because the loss can always be reduced by pushing the clusters further away from each other. This results in the clusters converging to exactly the set upper boundary value.
A second regularization loss based on the Kullback–Leibler divergence is applied to encourage balanced soft assignments of latent space data points to fuzzy rules, following the approach proposed in~\cite{Aljalbout_Golkov_Siddiqui_Strobel_Cremers_2018}.
%\begin{equation}
%    \mathcal{L}_{H} = \sum_i^c \sum_k \frac{\Psi_i^2(k)}{\sum_k\Psi_i^2(k)} %\log{\frac{\Psi_i^2(k)/\sum_k\Psi_i^2(k)}{\Psi_i(k)}}
%\end{equation}
\begin{equation}
    \mathcal{L}_{B} = \sum_i^c \frac{\sum_k\Psi_i(k)}{N} \log{\frac{{\sum_k\Psi_i(k)}/{N}}{1/c}}
\end{equation}
This balanced assignment loss encourages all clusters to have a uniform probability of being assigned a data point. %While this is not always desired \cite{Aljalbout_Golkov_Siddiqui_Strobel_Cremers_2018}, it is beneficial in our case as it encourages the latent space data to be in proximity (have a low distance $d_i^2(k)$) to all clusters instead of just one. %However, since the number of latent space data samples is based on the batch size, it affects the quality of the deep clustering. The batch size should be much larger than the number of clusters or fuzzy rules.
\section{Experimentation}\label{sec:3}
\par The proposed methodology was applied to the multi-horizon time series prediction of the closing prices of the Standard \& Poor's 500 stock market index. 
% Tracking the stock performance of the largest companies listed on U.S. stock exchanges, the S\&P 500 is regarded as a reliable indicator of the overall market. 
We examined multivariate data from other related market indicators: the VIX Volatility Index, a commonly referenced measure of the stock market's expectation of volatility based on S\&P 500 index options, often referred to as the "fear index"; the Gold commodity price; and the 5-year U.S. Treasury Yield, which tracks the performance of U.S. dollar-denominated domestic sovereign debt. 
Data were gathered from January 1st, 2001, to January 1st, 2023, then normalized using Min-Max scaling and split into training (80\%), validation (10\%), and testing (10\%) sets. 
% The data were not shuffled to preserve order, allowing for an examination of the trained model's performance on states not present in the training data. 
% The data for the main series (S\&P 500) falls within the ranges [$0$, $0.533$], [$0.379$, $0.705$], and [$0.629$, $1$] for the training, validation, and testing sets, respectively.

\par As a baseline we compared the Fuzzformer with the classic Autoregressive Integrated Moving Average (ARIMA) model and the Long Short-Term Memory (LSTM) Recurrent Neural Network (RNN) for time series forecasting\footnote{{https://github.com/mihaozbot/Fuzzy-transformer}}. 
% The ARIMA metaparameters were selected with the Auto ARIMA method\footnote{\url{https://github.com/robjhyndman/forecast/blob/master/R/arima.R}}, which resulted in ARIMA(p,d,q), where $p {=} 4$ is the number of autoregressive terms, $d {=}1$ is the order of differentiation for nonseasonal stationarity, and $q {=} 1$ is the number of lagged forecast errors. 
% The approach of training, validation, and testing is not applicable for the ARIMA model. 
The ARIMA meta-parameters were selected with the Auto ARIMA method.
The following parameters were used for the LSTM model: hidden dimensions were 32 and 128, and the number of layers was 3. 
A linear layer was also used to transform the output into $\mathbb{R}^{N\times 1}$, and a number of time steps equal to the number of forecast horizons were taken as the output of the model. 
The meta-parameters of the Fuzzformer network were selected experimentally as: 2 LSTM layers, 2 MHA layers, hidden dimension $D_h {=} 128$, latent space dimension $D_Z {=} 2$, number of attention heads $h {=} 4$, order of auto-regression  $p{=}2,4,30$, differentiating order $d{=}1$, exogenous input order $q{=}1$, number of fuzzy rules $C = 16$. 
% The batch size was $B = 313$ in all cases. 
The methods were evaluated based on the Root Mean Squared Error (RMSE) measure. Importantly, the ARIMA model is not directly comparable with deep learning methods, as it does not retain long term data, but estimates coefficients from each input data sequence.

% \begin{equation}
%  \mathrm{RMSE} =\sqrt{\frac{1}{M} \sum_{ k = 1}^M \left(\hat{\underline{y}}_k-\underline{y}_k\right)^2}
%  \end{equation}
% where $\hat{\underline{y}}_k$ is the future forecast at time step $k$, and $M$ is the size of the dataset.

\begin{table*}[!t]
\renewcommand{\arraystretch}{1}
\caption{Comparison of the proposed Fuzzformer neuro-fuzzy system with classic multi-horizon time series forecasting methods}
\label{tab:1}
\centering
{\footnotesize%
\begin{tabular}{ll | lll | lll | lll}%{|l|c|c|c|c|c|c|c|c|c|c|c|} 
\specialrule{.15em}{.05em}{.05em}
\multicolumn{2}{l|}{ $N$/$H$} & \multicolumn{3}{c|}{60/30} & \multicolumn{3}{c|}{150/30} & \multicolumn{3}{c}{150/60} \\
\multicolumn{2}{l|}{Dataset} & Train & Valid & Test & Train & Valid  & Test & Train & Valid  & Test\\
\specialrule{.15em}{.05em}{.05em}
\multicolumn{2}{l|}{Auto ARIMA($p {=} 4$, $d{=} 1$, $q{=} 1$)} & 0.0096 &  {0.0288}  &  0.0313 & 0.0091  &   {0.0290} &  {0.0320} & 0.0123  &   {0.0424} &  {0.0441}\\
\multicolumn{2}{l|}{ARIMA($p {=} 30$, $d{=} 1$, $q{=} 1$)} & 0.0091 &  0.0303 &  {0.0311} & 0.0100 & 0.0326 & 0.0355 & 0.0133 & 0.0472& 0.0462\\
\hline
\multicolumn{2}{l|}{LSTM($D_h{=} 256$, 3 layers)} &0.0124 &   \bred{0.0407} &  \bred{0.1436}  & {0.0128} &  \bred{0.0416} & \bred{0.1536} & 0.0163 & {0.0492}  & \bred{0.1683}\\
\multicolumn{2}{l|}{LSTM($D_h{=} 256$, 1 layer)} &  0.0121  & 0.0386  & 0.0797 & {0.0118} & {0.0395}  & 0.0981 & 0.0162 &  \bred{0.0549}  & 0.1181\\
\multicolumn{2}{l|}{Fuzzformer (Our, $p {=} 2$)} & 0.0100 & 0.0303 & 0.0324 & 0.0098 & 0.0308 & \bgreen{0.0333} & 0.0128 & \bgreen{0.0428}& 0.0452\\
\multicolumn{2}{l|}{Fuzzformer (Our, $p {=} 4$)} & 0.0107 & 0.0320 &  0.0369 & 0.0098 & 0.0314  & 0.0338 & 0.0183 & {0.0450} &  {0.0462} \\
\multicolumn{2}{l|}{Fuzzformer (Our, $p {=} 30$)} & {0.0095} & \bgreen{0.0300} & \bgreen{0.0321} & {0.0094} &  \bgreen{0.0301}  & {0.0336} & 0.0130 &  0.0455 & \bgreen{0.0442}\\
\specialrule{.15em}{.05em}{.05em}
\end{tabular}
}%
\end{table*}
\par
\begin{figure}[!t]
  \centering

  % Top row
  \begin{minipage}{0.395\linewidth}
    \centering
    \includegraphics[width=\linewidth]{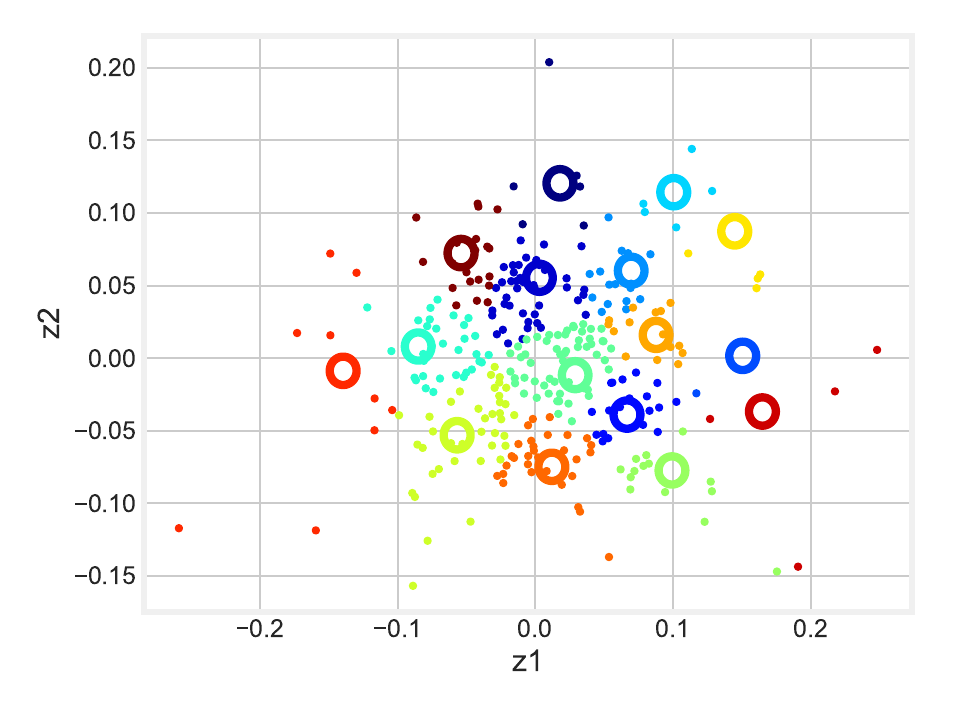}
    {\footnotesize (a) Antecedent clusters}
  \end{minipage}
  \hfill
  \begin{minipage}{0.585\linewidth}
    \centering
    \includegraphics[width=\linewidth]{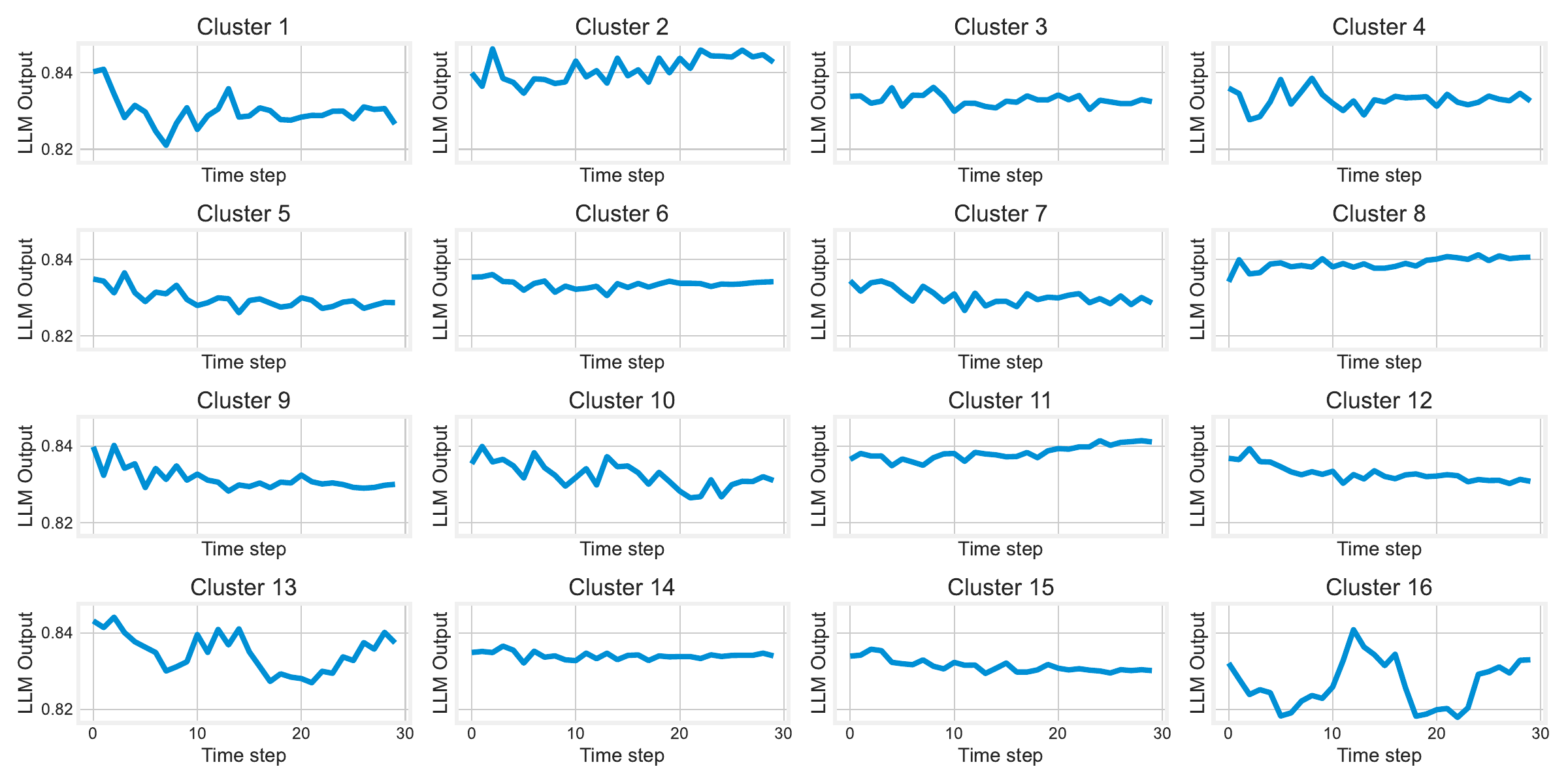}
    {\footnotesize (b) Consequence ARIX models} 
  \end{minipage}

  \vspace{1.5ex} % space between rows

  % Bottom image
  \begin{minipage}{1\linewidth}
    \centering
    \includegraphics[width=\linewidth]{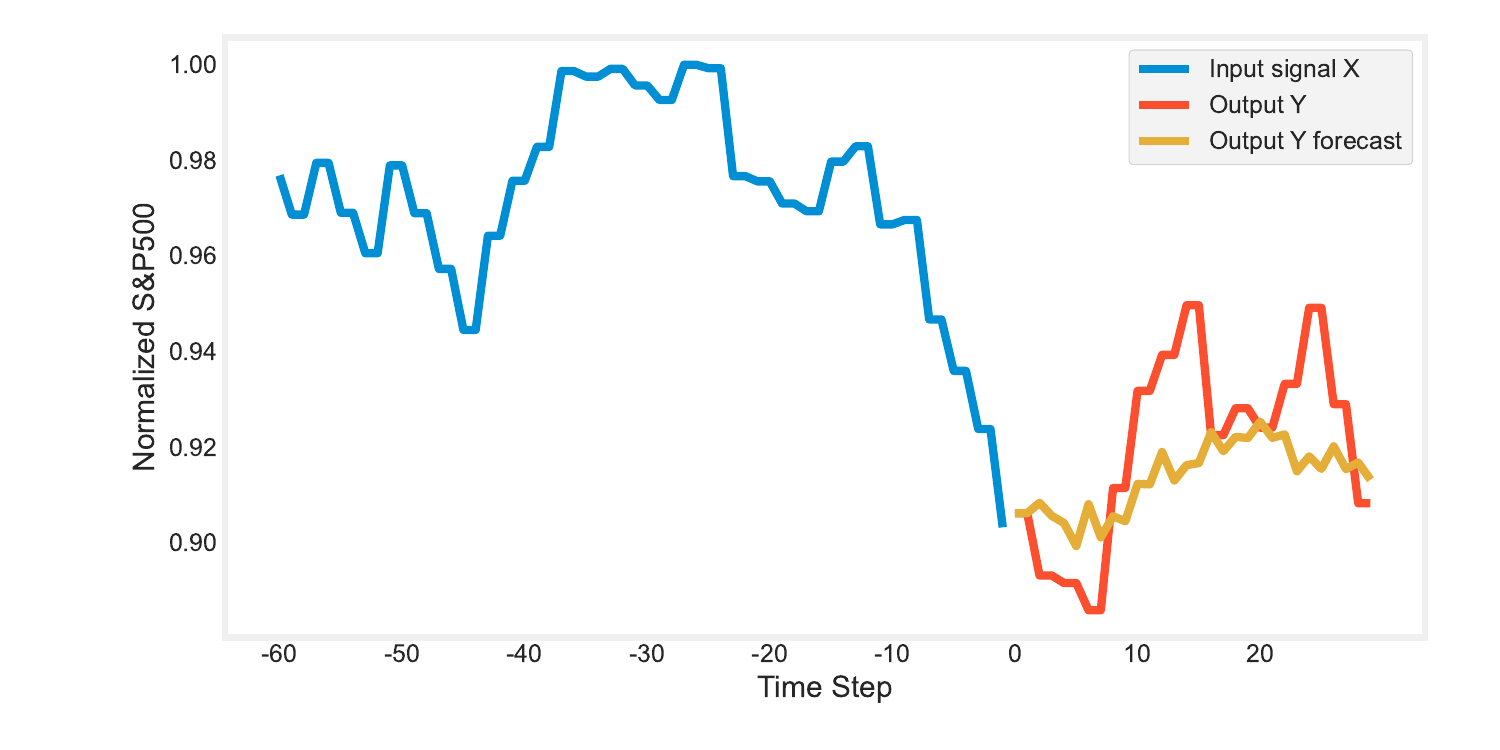}
    {\footnotesize (c) Combined output}
  \end{minipage}

  \caption{A multi-horizon forecast of the Fuzzformer with $p = 30$, test length $N {=} 60$, and horizon $H {=} 30$. The model predicts a bounce after a market drop.}
  \label{fig:5}
\end{figure}

 The results of the case study, presented in Table \ref{tab:1}, demonstrate that the Fuzzformer model is less prone to overfitting than the LSTM model when tested on the given data. An example of the prediction is presented in Fig.~\ref{fig:5}. 

% \begin{figure}[!t]
% \centering
%   \includegraphics[width=0.6\linewidth]{Images/fig.3.pdf}
%     \caption{The neural network latent space $\underline{Z}{\in}\R^2$ with the fuzzy model multivariate Gaussian clusters and corresponding highest activation of membership functions for one batch of data with 313 samples. The membership functions of the fuzzy rules provide a label to the latent space data samples. \label{fig:3}
%     }%
% \end{figure}
% % \begin{figure}[!t]
% \centering
%   \includegraphics[width=1\linewidth]{Images/fig.4.pdf}
%     \caption{The forecasts of the local ARIX models $p{=}30$ of the fuzzy layer for one input sequence with $N {=} 60$ and $H {=} 30$. The final output is computed as a fuzzy combination of all local models. \label{fig:4}
%     }%
% \end{figure}
% The Fuzzformer model provides more conservative forecasts, producing nearly linear slopes that mainly describe the future trend. 
In contrast, the LSTM network exhibits higher variance in predictions and, as a consequence, much larger prediction errors on the test data. 
Though the LSTM network performs well on the training data, the quality of its forecasts declines sharply on the test dataset, an indication of overfitting. 
% This susceptibility to the amplitude ranges in the datasets, as set up in the experiments, likely contributes to this issue. 
% Interestingly, more layers in the case of the LSTM network do not reduce the forecast error. 
Conversely, the Fuzzformer model does not suffer from this problem, thanks to the use of simple ARIX local models, but it requires more epochs to train effectively due to the attention layer and individual recursive computation for each rule, making training slower. 
From the second column in Table \ref{tab:1}, it's apparent that extending the look-back period to $N{=}150$ yields no benefits for either the LSTM or Fuzzformer networks.

\par The ARIMA models with $p=4$ generally forecast a constant value without a trend, except when a pronounced trend is present over the entire look-back window. 
This approach results in forecasts with low RMSE, as the error is never too substantial. 
% In comparison, the Fuzzformer model produces cautiously optimistic predictions, revealing trends but avoiding the extremes observed with ARIMA, even when a clear trajectory exists in the look-back window. 
% This behavior may be attributable to the training process, where continuing a downward trend leads to large RMSE errors if the series reverses direction (bounces). 
The ARIMA model of higher orders may have some decomposition singularity issues during identification. 
However, our proposed ARIX model, which uses a similar autoregressive structure and is trained with the Adam optimization method, can use much higher orders without difficulty. 
Training an autoregressive model with a gradient descent method is more stable for higher system orders. 
% Still, it's worth noting that the trained parameters might be biased in the classical system identification sense. 
The use of the neural network encoder allows the fuzzy system head to use multivariate data, as AR/ARX/ARIX/ARIMAX type models can handle only univariate data. 
The fuzzy model combines multiple local models in a manner similar to how samples are distributed and multiple models in a SARIMAX model, but with fuzzy model membership based on antecedent clusters.

\section{Conclusion}\label{sec:5}
\par In this study, we examined how a recurrent neural network architecture with multi-head self-attention can condense multivariate stock market data into features that can be utilized for an interpretable fuzzy multi-horizon time series forecast. 
% The method enables the incorporation of multivariate data, thereby enhancing the information available as input to the fuzzy inference systems, which typically have lower descriptive power and encounter difficulties with multivariate data.
The proposed approach demonstrates comparable performance with established ARIMA and LSTM networks that are commonly used for multi-horizon time series forecasting while offering valuable insights into the model structure and information flow through the network.
% The temporal attention mechanism highlights the significance of delayed time steps, and the fuzzy inference system offers interpretability with the cluster-based membership functions and simple consequent models.
Nonetheless, there remain opportunities for incorporating further interpretable mechanisms into the proposed fuzzy system that were not implemented in this study.
% The initialization of clusters could also be enhanced with an initial clustering method, and the dependency of clustering on the batch number is a limitation to be addressed.
These initial results are promising, and with continued research, they could contribute to a deeper understanding of stock market behavior.

%The evsystem would provide interpretability and model adaptation in a changing environment, defined by abrupt conceptual shifts and slow drifts that are commonly present in the stock market.
\bibliographystyle{IEEEtran}
%\bibliography{./references.bib}

%

%
\end{document}